\documentclass[10pt,twocolumn,letterpaper]{article}

\usepackage{wacv}              
\usepackage{times}
\usepackage{epsfig}
\usepackage{graphicx}
\usepackage{amsmath}
\usepackage{amssymb}
\usepackage{booktabs}
\usepackage{float}
\usepackage[accsupp]{axessibility}
\usepackage[pagebackref,breaklinks,colorlinks]{hyperref}

\usepackage[capitalize]{cleveref}
\crefname{section}{Sec.}{Secs.}
\Crefname{section}{Section}{Sections}
\Crefname{table}{Table}{Tables}
\crefname{table}{Tab.}{Tabs.}

\usepackage[nolist,nohyperlinks]{acronym}
\acrodef{DL}{Deep learning}
\acrodef{DA}{Domain Adaptation}
\acrodef{CNN}{Convolutional Neural Network}
\acrodef{SOTA}{state-of-the-art}
\acrodef{AI}{Artificial Intelligence}
\acrodef{ML}{Machine Learning} 
\acrodef{DG}{Domain Generalization}

\def\wacvPaperID{648} 
\def\confName{WACV}
\def\confYear{2024}

\begin{document}

\title{Domain Generalization with Correlated Style Uncertainty}

\author{Zheyuan Zhang$^{1}$,
Bin Wang$^{1}$,
Debesh Jha$^{1}$,
Ugur Demir$^{1}$,
Ulas Bagci$^{1}$ \\
$^{1}$ Machine \& Hybrid Intelligence Lab, Northwestern University, USA\\
{\tt\small ulas.bagci@northwestern.edu}
}
\maketitle

\begin{abstract}
   Domain generalization (DG) approaches intend to extract domain invariant features that can lead to a more robust deep learning model. In this regard, style augmentation is a strong DG method taking advantage of instance-specific feature statistics containing informative style characteristics to synthetic novel domains. While it is one of the state-of-the-art methods, prior works on style augmentation have either disregarded the interdependence amongst distinct feature channels or have solely constrained style augmentation to linear interpolation. To address these research gaps, in this work, we introduce a novel augmentation approach, named \textit{Correlated Style Uncertainty (CSU)}, surpassing the limitations of linear interpolation in style statistic space and simultaneously preserving vital correlation information. Our method's efficacy is established through extensive experimentation on diverse cross-domain computer vision and medical imaging classification tasks: PACS, Office-Home, and Camelyon17 datasets, and the Duke-Market1501 instance retrieval task. The results showcase a remarkable improvement margin over existing state-of-the-art techniques. The source code is available \href{https://github.com/freshman97/CSU}{https://github.com/freshman97/CSU}.
\end{abstract}

\vspace{-6mm}
\section{Introduction}
\vspace{-1mm}
\label{sec:intro}
Recent years have witnessed the remarkable success of \acf{DL} in computer vision 
domain operating under the premise that the training (source) and testing (target) datasets adhere to a principle of independent and identically distributed (iid) data~\cite{zhou2022domainreview}. However, this oversimplified assumption often fails in practice when there is a distribution drift between training and testing datasets. The violation of this assumption induces the phenomenon that well-trained model in the source domain degrades dramatically in the target domain. If domain generalization is successfully implemented within a \ac{DL} model, it would inherently solve issues related to domain shift. This would not only signify a significant advancement in the field, but also streamline the practical deployment of DL models across various domains. For example, a car detector should perform accurately both on sunny and cloudy days. DL based medical image segmentation algorithm, for another example, should generate stable segmentation regardless of the acquisition and scanner differences, and so on. 

\begin{figure}[!t]
  \centering
  \includegraphics[width=0.8 \linewidth]{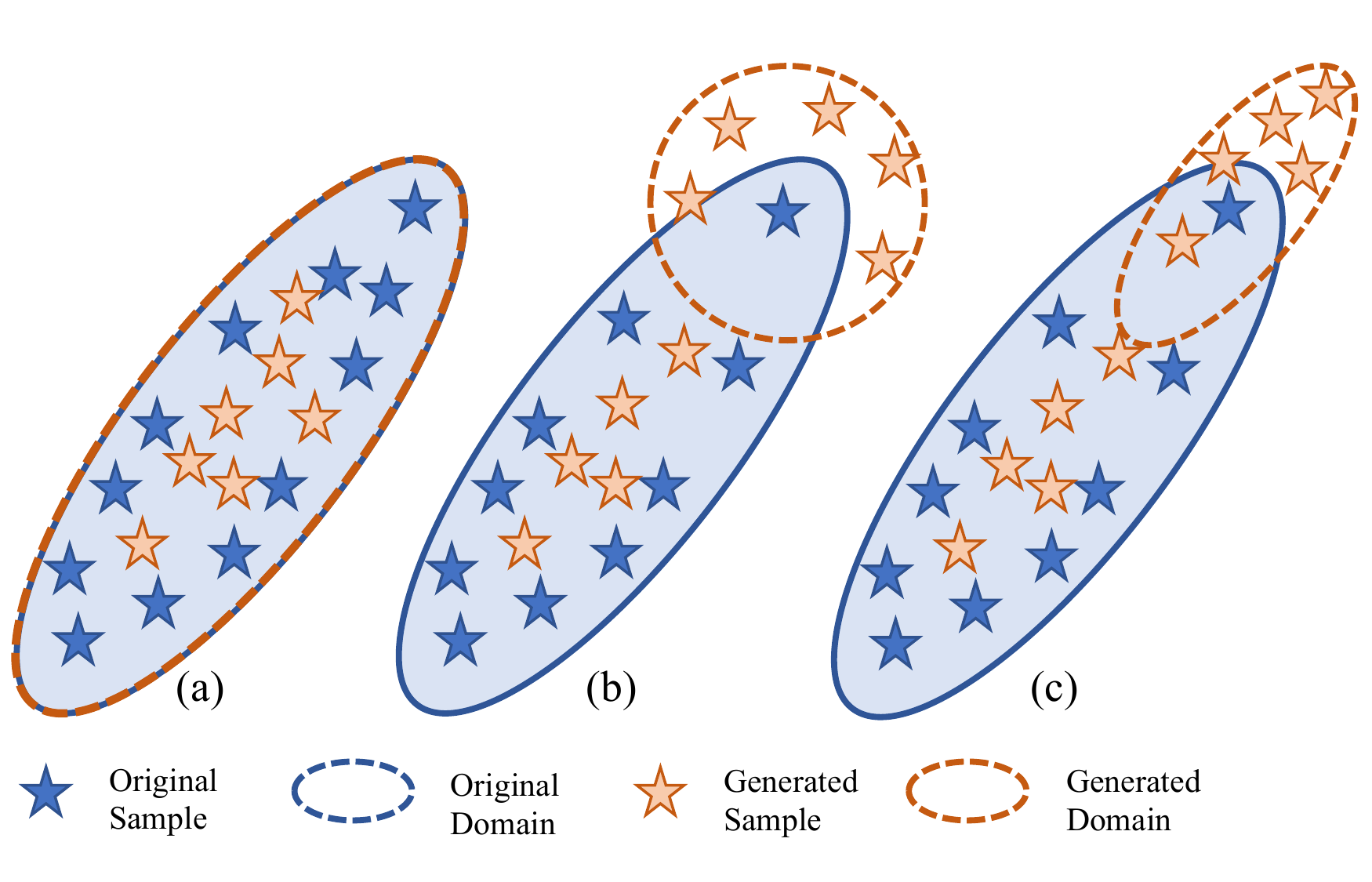}
   \caption{Visualization of synthetic feature statistics samples using (a) MixStyle~\cite{zhou2021mixstyle}, (b) DSU ~\cite{li2022dsu}, and (c) Our proposed Correlated Style Uncertainty (CSU) method. CSU preserves the correlation among feature channels.}
   \label{fig:illu}
   \vspace{-7 mm}
\end{figure}

\textbf{Domain adaptation} and \textbf{domain generalization} are two distinct approaches for addressing the challenge of domain shift in machine learning. A widely adopted approach to counteract domain shift problems entails acquiring unlabeled data from the target domain. By leveraging this data, we can adapt a model, initially trained on the source domain, to align better with the characteristics of the target domain. This strategy is called \acf{DA}, which has been the subject of much systematic investigation in the last few years and achieved promising results in many fields~\cite{zhou2022domainreview}. To ensure robust model adaptation, regularization algorithms such as entropy regularization~\cite{saito2019semientropy, prabhu2021sentryselectentro} are also applied to target domain during training. Domain adaptation assumes that the target domain is known and that some data from the target domain is available for training. However, accessing the target domain data can be quite challenging. Specifically in high-risk applications (e.g., medical data), target domain data might not be available at all. This is a primary research concern of \acf{DG} as a strong alternative path to \ac{DA}. 

In \ac{DG}, the objective is to develop models that can perform well across a wide range of domains without explicit training on each individual domain. Unlike \ac{DA}, \ac{DG} assumes that the target domain is unknown and that no data from the target domain is available for training~\cite{wang2022generalizingreview, zhou2022domainreview}. Since target data is unavailable for investigating domain shift, \ac{DG} relies solely on extracting robust, domain-invariant feature representations from diverse training distributions. As a result, when domain information is feasible, feature alignments among different domains could significantly leverage the model's out-of-distribution generalization ability, as shown in previous research~\cite{li2020domainalign, lu2022domainalign}. 

\textbf{What do we propose?} In this paper, we propose a novel DG method, called \textit{Correlated Style Uncertainty (CSU)}. The method is specifically designed to retain the relationship among distinct feature space channels while addressing the distribution drift between target and source domains. The effectiveness of our approach, as depicted in Figure \ref{fig:illu}(c), is clearly evident when compared with widely-used techniques like MixStyle and DSU. Like MixStyle and DSU, our newly introduced algorithm, CSU, belongs to the family of \textit{style augmentation methods} employed for DG, but it presents distinctive advantages. 

\textbf{How is CSU different from previous style augmentation?} In the context of style augmentation, MixStyle \cite{zhou2021mixstyle} operates by randomly choosing two samples and performing linear interpolation between their corresponding style attributes, to put it succinctly. While MixStyle is useful, it's confined to generating in-distribution samples, potentially limiting network's generalization capabilities. pAdaIN \cite{nuriel2021pAdaIN} rearranges the instance-specific feature statistics within a batch, thus sharing the same problem with MixStyle. DSU \cite{li2022dsu}, contrarily, addresses this limitation by creating out-of-distribution samples via uncertainty modeling of feature statistics. However, DSU's effectiveness is hinged upon the assumption that each channel operates independently, implying that inter-channel correlations bear no impact on the task at hand. This assumption might not always hold in practical scenarios. Style Neophile \cite{kang2022stylen} select style prototypes that represent the distribution of source styles stored in the source style queue using MMD distance from style storage during training. \cite{zhang2023adverstyle,zhong2022adverstyleseg} apply the adversarial attack on the feature statistics for domain generalization while introducing expensive computational burden.

Our proposed method, CSU, is fundamentally different from these methods because i) it produces out-of-distribution samples, ii) correlation among different channels are retained and iii) the sampling process requires a negligible computational burden. Our proposed method, CSU, allows us to generate more meaningful style perturbations in discriminatory tasks, thus enhancing the model's generalization capacity. Similar to hypotheses in prior research in the literature, we posit that the feature statistics follow a multivariate Gaussian distribution; however, we acknowledge and account for correlations among the variates, unlike many traditional methods. This perspective allows us to consider a more realistic distribution that could potentially improve model performance in real-world scenarios.

In our specific approach, we first compute the covariance matrix at the mini-batch level and then estimate the distribution from the covariance matrix.  This allows us to sample correlated feature statistics from the established distribution. This sampling allows us to generate the style statistics outside the linear interpolation while maintaining an identical correlation. In this way, more diverse but meaningful style augmentation can be applied during the training and increase the model's generalization ability. We highlight our main contributions in this study as follows:
\begin{itemize}
\item Our proposed strategy (CSU) is a well-calibrated method that goes beyond the interpolation strategies by preserving correlation between different feature spaces. This allows us to generate more diverse and meaningful style augmentation during training which helps in building a more generalizable model. To the best of our knowledge, such a simple yet effective style augmentation strategy has never been explored. 
    
\item To evaluate the effectiveness of the proposed CSU model, we conducted extensive experiments on multi-domain classification benchmarking datasets, including PACS~\cite{li2017deeperPACS}, Office-Home\cite{venkateswara2017deepofficehome}, Camelyon17~\cite{bandi2018detectioncamelyon} and the Duke-Market1501 dataset for instance retrieval tasks~\cite{ristani2016performanceDuke, zheng2015scalablemarket}. The quantitative results show that the CSU can significantly improve the model's generalizability over other \acf{SOTA} methods.
    
\item We have performed several ablation studies to investigate the optimal parameters for the CSU model. These investigations have covered factors such as the ideal position for model integration, the most efficient sampling hyperparameters, and the batch size that yields the greatest level of generalization.
\end{itemize}
\vspace{-3.5mm}
\section{Background and Related Works}

Several \ac{DG} strategies have been proposed in the literature;  we briefly cover the mostly used ones under sub-categories as follows.

\textbf{Data Augmentation:}
is highly valued for its role in exposing the model to an expanded set of instances, an essential element for successful Deep Learning. Many methods have been proposed to achieve strong data augmentation, including traditional image augmentation like BigAug~\cite{zhang2020generalizingBigAug}, deep neural network-based image generation like in RandConv~\cite{xu2020robustrandconv}, and adversarial data augmentations \cite{volpi2018generalizingadversarialaug}. These methods are  suitable specifically when the domain tags of samples are agnostic. While data augmentation is a powerful technique that can improve the generalization performance of machine learning models, it also has some potential drawbacks in the context of domain generalization. For instance, one drawback is that data augmentation may not be effective when the variations between the source and target domains are too significant. \textbf{Feature Alignment:}
is a popular method in representation learning category of DG approaches. Given domain tags, the model will add regularization terms into loss functions to force the extracted features from all source domains to align to the same distribution. For instance, Li~\cite{li2018domainmaxmeandiscrepany} introduced the Maximum Mean Discrepancy as a regularization term to achieve feature alignment across multiple domains. Zhao~\cite{zhao2020domainentropy} proposed an entropy regularization term that measures the dependency between the learned features and corresponding labels. This regularization method can ensure the conditional invariance of learned features. One major drawback of feature alignment  is that it can be difficult to determine the best way to align the features across domains. \textbf{Meta-learning:}
has also attracted attention from DG communities~\cite{choi2021meta, shu2021openmeta}. Meta-learning aims to learn the learning algorithm itself by learning from previous experience or tasks. By splitting the source domain samples into pseudo-train and pseudo-test, meta-learning mimics the potential domain shift of the actual target domain. Despite its promise, meta-learning can be computationally expensive and time-consuming. Since meta-learning involves training a model on a large number of tasks or domains, there is a risk that the model may overfit to the training data and not generalize well to new domains.  

\textbf{Style Augmentation:}
The final category of DG is very recent: \textit{style augmentation}. This method comes from the simple observation that instance-specific feature statistics such as mean and standard deviation, contain informative style characteristics and can be applied to the style-transferring model~\cite{huang2017arbitrarystyle}. This phenomenon allows us to generate different style images while maintaining the same semantic concept. For example, Seo et al.~\cite{seo2020learningDOSN} proposed one domain-specific normalization method by calculating the feature statistics of each domain.  Pan  et al.~\cite{li2021sfa} add perturbation on the feature embedding with simple Gaussian noise during training.  Zhou et al.~\cite{zhou2021mixstyle} presented mixing styles (MixStyle) of training instances, and increased the source domain diversity. As a result, authors leveraged the trained model's generalizability. Nuriel et al.~\cite{nuriel2021pAdaIN} alternatively proposed a Permuted Adaptive Instance Normalization (pAdaIN) method to rearrange the instance-specific feature statistics within a batch, thus improving the model's generalizability. In a slightly different angle, Li et al.~\cite{li2022dsu} quantified feature statistics' uncertainty (DSU) and sampled new style feature statistics from the uncertainty distribution,  resulting in novel out-of-distribution domains being synthesized implicitly. Kang et al.~\cite{kang2022stylen} select style prototypes from style storage during training using the MMD distance and these prototypes represent the distribution of source styles stored in the source style queue. \cite{zhang2023adverstyle,zhong2022adverstyleseg} apply the adversarial attack on the feature statistics for domain generalization while introducing expensive computational burden. Our work is directly related to MixStyle and DSU as we are using the same hypothesis but addressing the major limitations of them, and sharing some similarities with pAdaIN, from the same efforts for synthesizing novel domains. Unlike them,  our proposed CSU generates out-of-distribution feature statistics while maintaining the correlation between features.

\vspace{-1mm}
\section{Methods}
\vspace{-1mm}
\subsection{Correlation within the style statistics}
\begin{figure}[!h]
  \centering
  \includegraphics[width=1.0\linewidth]{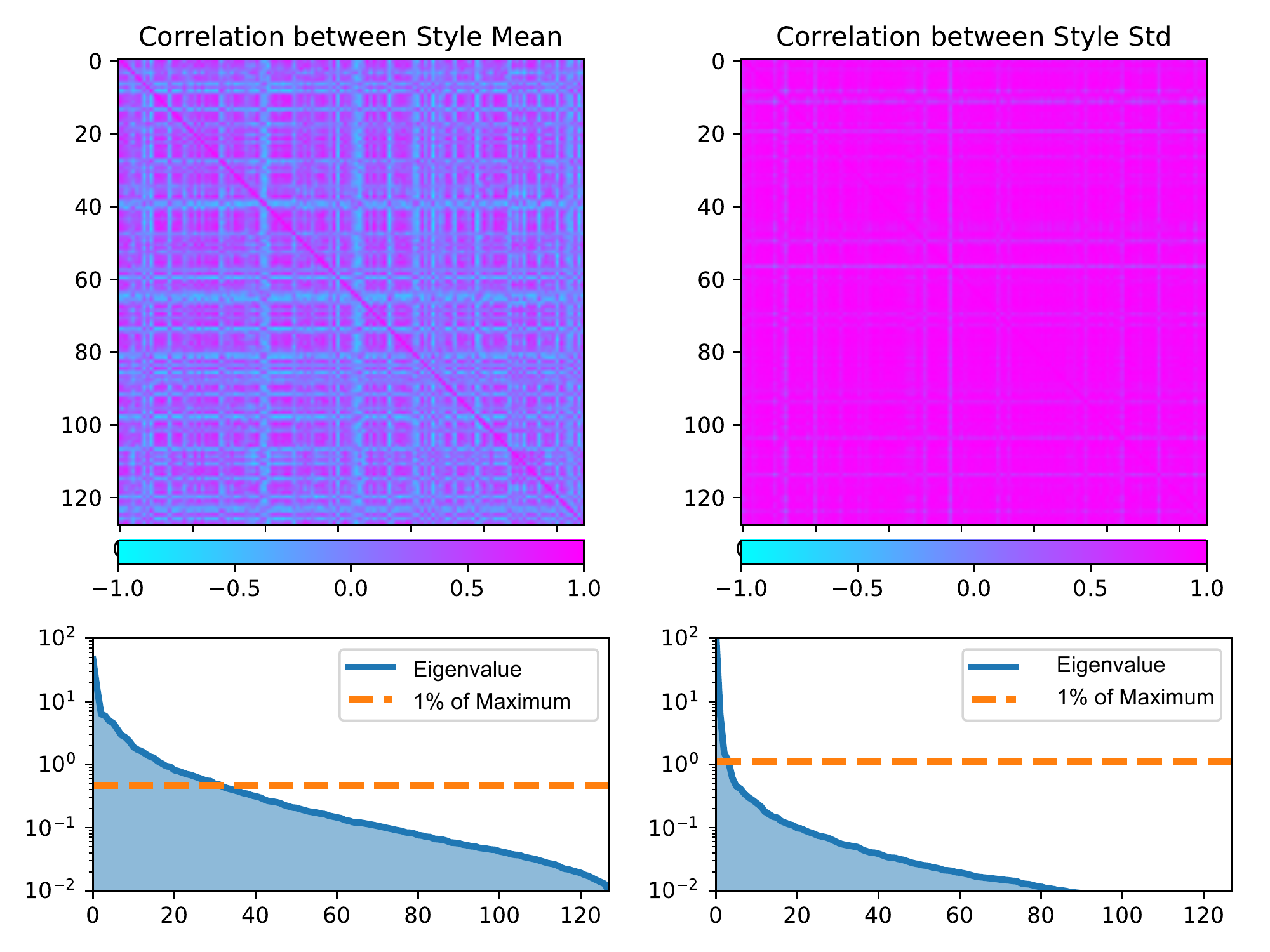}
   \caption{Visualization of feature statistics correlation. We calculate the style statistics (mean and standard deviation, respectively) on the PACS dataset. We extract the feature using the second residual block output from the ImageNet pretrained on ResNet18~\cite{he2016deepresnet} with a channel size of 128. For 4 domains of the PACS dataset, including Art, Cartoon, Photo, and Sketch, we select 64 cases from every category (7 categories in total) under each domain. Therefore, the data samples to calculate the correlation matrix is $7\times4\times 64=1792 >> 128$.}
   \label{fig:correlation}
   \vspace{-5mm}
\end{figure}

\noindent Given batch level feature maps $x \in \mathbb{R}^{B\times C \times H \times W}$ of the network $f(in, \phi)$ where $in$ denotes the batch-wise inputs and $\phi$ denotes the network parameters. We can formulate the instance-specific feature statistics mean $\mu \in \mathbb{R}^{B\times C }$ and standard deviation $\sigma \in \mathbb{R}^{B\times C }$ as follows
\begin{gather}
    \mu(x) = \frac{1}{HW} \sum_{h=1}^{H} \sum_{w=1}^{W} x_{b, c, h, w}, \\
    \sigma^2(x) = \frac{1}{HW} \sum_{h=1}^{H} \sum_{w=1}^{W} (x_{b, c, h, w}-\mu(x))^2.
    \label{eq:mu sigma}
\end{gather}
Thus, we can formulate the channel-wise covariance matrix $\Sigma_{\mu} \in \mathbb{R}^{C \times C }$, $\Sigma_{\sigma} \in \mathbb{R}^{C \times C }$ of $\mu, \sigma$:

\begin{gather}
    \Sigma_{\mu} = \frac{1}{B} (\mu -E(\mu))^T (\mu -E(\mu)), \\
    \Sigma_{\sigma} = \frac{1}{B} (\sigma -E(\sigma))^T (\sigma -E(\sigma)),
    \label{eq:Sigma mu sigma}
\end{gather}
where $E(\mu), E(\sigma)$ represents the mean value of $\mu, \sigma$ over batch dimension. It is worth noting that the rank of $\Sigma_{\mu}, \Sigma_{\sigma}$ is strictly limited by $min(B, C)\leqslant	C$. Previous research indicated that the feature maps are unlikely to be linearly independent over the channel dimension~\cite{zhang2021stochasticwbn, huang2018decorrelatedbn}. 
Without any form of regularization, it becomes difficult to assume a diagonal covariance matrix and a zero correlation across each channel, as suggested in the top row of Figure~\ref{fig:correlation}.

\begin{figure*}[h!]
  \centering
  \includegraphics[width=.8 \linewidth]{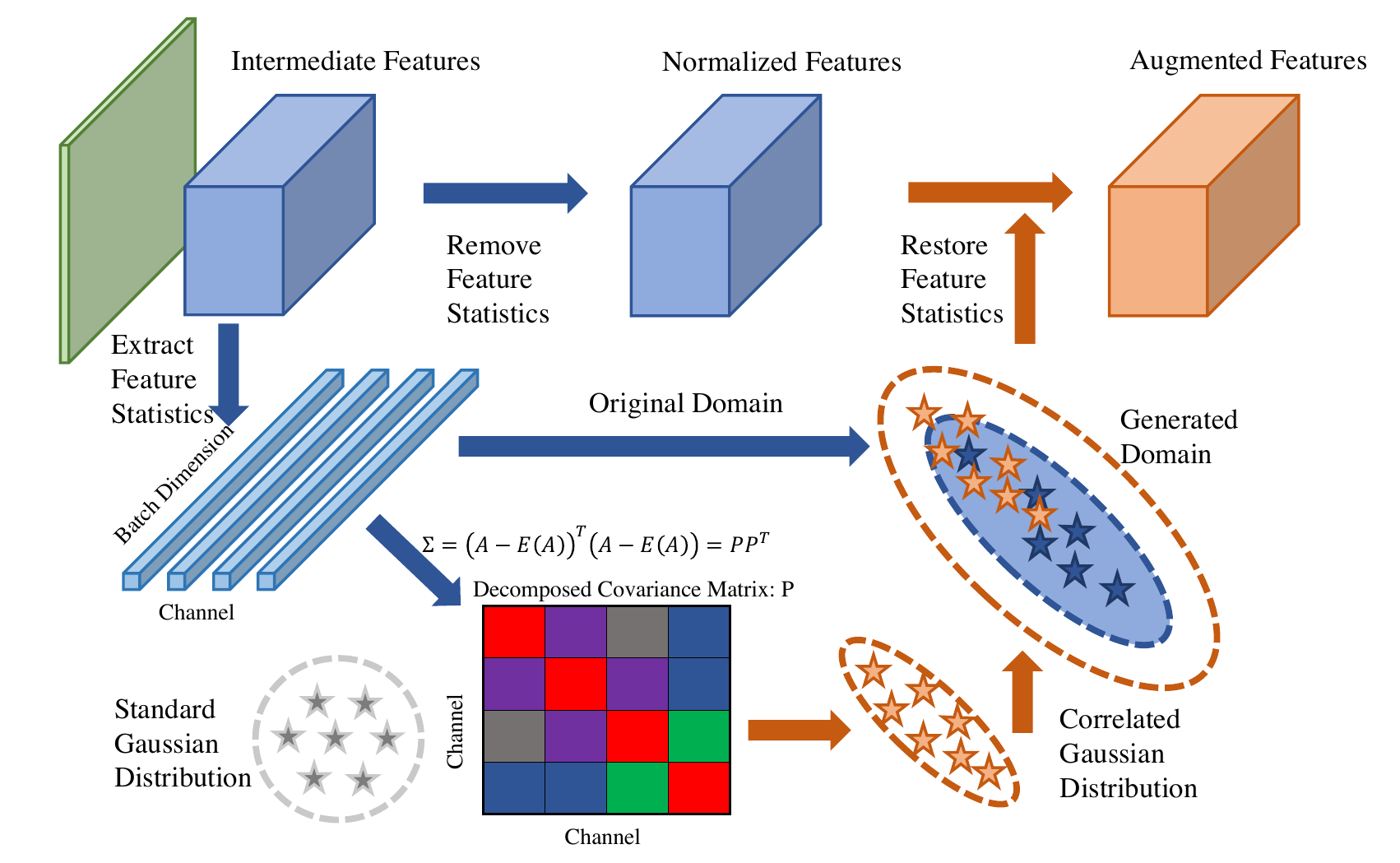}
   \caption{Visualization of feature statistics augmentation using correlated style uncertainty (CSU). Given the intermediate features extracted from the network, we first estimate the covariance matrix of feature statistics and decompose the covariance matrix as described in Sec~\ref{sec:model}. Based on this decomposition, we could generate correlated augmentation from the standard Gaussian distribution that shares identical distribution as the original domain. Then, we update raw data feature statistics by adding this correlated augmentation. Finally, we restore the feature statistics back to the normalized features and achieve the augmented features.}
   \label{fig:network}
   \vspace{-5 mm}
\end{figure*}
We observed that the correlation matrix of style statistics (regardless of the mean or the standard deviation values) is not diagonal. Indeed, there exists a strong correlation among the channels. We applied eigenvalue decomposition over the calculated correlation matrix and find that very few eigenvectors dominate most variance, as shown in the bottom row of the Figure~\ref{fig:correlation}. This inspired us to rethink the augmentation of style statistics. The correlation matrix indicates that the combinations of feature statistics are not arbitrary but limited by task objectives and training procedures. Most variances happen within the specific principal directions. Arbitrary augmentation over the style statistics might damage the training itself. Previous research about why InstanceNorm can not outperform the BatchNorm in the discriminative tasks also proves this finding~\cite{nam2018batchinstancenorm}.

In reevaluating feature statistics augmentation, we observe that both MixStyle~\cite{zhou2021mixstyle} and pAdaIN~\cite{nuriel2021pAdaIN}  limit their augmentations within the original feature space, preserving either channel information or channel combinations respectively. DSU~\cite{li2022dsu} attempts to exceed this limitation via uncertainty quantification, but hinges on strict orthogonality between channel feature statistics. Our proposed model, CSU, breaks these boundaries by generating feature statistics beyond the training domain while preserving channel correlations, offering a more robust feature augmentation. CSU builds upon the strengths of these strategies and addresses their limitations, significantly advancing the field of feature statistics augmentation. There are some other style augmentation methods like AdverStyle \cite{zhang2023adverstyle} or Style Neophile \cite{kang2022stylen}. However, direct mathematical correlation relationship analysis for these methods is not feasible due to complex adversarial training or distribution selection.

\vspace{-1mm}
\subsection{Modeling correlated style uncertainty}
\vspace{-1mm}
\label{sec:model}
Given that the correlation matrix is real, symmetric, and positive semi-definite, then we can apply eigenvalue decomposition on $\Sigma_{\mu}, \Sigma_{\sigma}$ to analyze its subspaces as:
\begin{gather}
    \Sigma_{\mu} = Q_{\mu} diag(\Lambda_{\mu}) Q_{\mu}^T, \\
    \Sigma_{\sigma} = Q_{\sigma} diag(\Lambda_{\sigma}) Q_{\sigma}^T,\\
    Q_{\mu} Q_{\mu}^T = Q_{\sigma} Q_{\sigma}^T = I, \\
    \Lambda_{\mu}, \Lambda_{\sigma} \in \mathbb{R}^{C}, Q_{\mu}, Q_{\sigma} \in \mathbb{R}^{C \times C },
    \label{eq:Decomposition}
\end{gather}
where $\Lambda_{\mu, i} \geqslant \Lambda_{\mu, j} \geqslant 0$ , $\Lambda_{\sigma, i} \geqslant \Lambda_{\sigma, j} \geqslant 0$ ($i>j$) represent the sorted eigenvalues, and $Q_{\mu}, Q_{\sigma}$ are the corresponding eigenvectors. The eigenvector corresponding to the largest eigenvalue represents the direction that we could apply dense augmentation. Eigenvectors corresponding to the eigenvalues of 0 or close to 0 are not considered in the data augmentation process due to low variance across such directions within the dataset.

Assuming that the $\mu, \sigma$ still follows the multi-variable Gaussian distribution, and $k_{\mu}, k_{\sigma}$ represent the independent variable number (or the rank of the corresponding covariance matrix), then we can represent the probability distribution function as:
\begin{gather}
    f_{\mu} = \frac{1}{(2 \pi)^{k_{\mu}} \det^*(\Sigma_{\mu})} \exp^{-(\mu-E(\mu))^T \Sigma_{\mu}^+ (\mu-E(\mu))}, \\
    f_{\sigma} = \frac{1}{(2 \pi)^{k_{\sigma}} \det^*(\Sigma_{\sigma})} \exp^{-(\sigma-E(\sigma))^T \Sigma_{\sigma}^+ (\sigma-E(\sigma))},
\end{gather}
where the $det^*$ is the pseudo-determinant and $\Sigma^+$ is the generalized inverse. Based on this distribution function, we further derive the correlated uncertainty augmentation after we sample $\epsilon_{\mu}, \epsilon_{\sigma} \in \mathbb{R}^{N \times C}$ from the standard Gaussian distribution $Y \sim \mathcal{N}(0,I)$ as follow:
\begin{gather}
    P_{\mu} = Q_{\mu} diag(\Lambda_\mu)^\frac{1}{2} Q_{\mu}^T, \\
    P_{\sigma} = Q_{\sigma} diag(\Lambda_\sigma) ^\frac{1}{2} Q_{\sigma}^T,\\
    \hat{\epsilon}_{\mu} = \epsilon_{\mu} P_{\mu}, \quad \hat{\epsilon}_{\mu}  = \epsilon_{\sigma} P_{\sigma}.
\end{gather}
Essentially, we determine the transform matrix $P$ such that the covariance matrix $ \Sigma = PP^T$ as shown in Eq.11-12, and these transformation matrices allow us to maintain the correlation. We sample the correlated perturbations $\hat{\epsilon} \in \mathbb{R}^{N \times C }$ from independent Gaussian noise $\epsilon \in \mathbb{R}^{N \times C }$ as shown in Eq.13. Note the covariance matrix:
\begin{gather*}
    \Sigma = PP^T = Q diag(\Lambda_\mu)^\frac{1}{2} Q^T (Q diag(\Lambda_\mu)^\frac{1}{2} Q^T)^T \\
    \quad = Q diag(\Lambda_\mu)^\frac{1}{2} (Q diag(\Lambda_\mu)^\frac{1}{2})^T.
\end{gather*}
These two transformations $Q diag(\Lambda_\mu)^\frac{1}{2} Q^T$, $Q diag(\Lambda_\mu)^\frac{1}{2}$ both work in principle. However, in practice, if we use the standard format of eigen-decomposition, stochastic axis swapping in the eigenvector will not influence this decomposition, but it can lead to unstable training. Thus, we apply the decomposition trick in the format of $Q diag(\Lambda_\mu)^\frac{1}{2} Q^T$ rather than using only one set of eigenvectors $Q diag(\Lambda_\mu)^\frac{1}{2}$ to avoid the random flip issue in traditional eigenvalue decomposition ~\cite{huang2018decorrelatedbn}. While we derive $\epsilon_{\mu}, \epsilon_{\sigma}$ from C independent normal variables, the resulting $\hat{\epsilon}{\mu}, \hat{\epsilon}{\mu}$  contain only $k_{\mu}, k_{\sigma}$ independent components, corresponding to the non-zero components of eigenvalues.
It is also worth noticing that the computation of eigenvalue decomposition is conducted by the torch operator with a complexity of $\sim O(N^3)$ in practice~\cite{strang2006linear}. Given the computation is conducted in polynomial time complexity and $N<512$ for most cases, the the extra time required is negligible.

\subsection{Style augmentation with CSU}
\vspace{-1mm}
Based on the previous two sections, we now present the style augmentation with correlated style uncertainty as follows:
\begin{gather}
    \beta (x) = \mu(x) + \lambda*\hat{\epsilon}_{\mu}, \\
    \gamma (x) =  \sigma(x) + \lambda*\hat{\epsilon}_{\sigma},
\end{gather}
where $\lambda\sim Beta(\alpha, \alpha)$ represents the augmentation intensity generated from the Beta distribution. Hyperparameter $\alpha$ controls the shape of the distribution. In the ablation experiments, we further show the influence of hyperparameter selections on the final performance. We can understand the equation in one more intuitive way, the first part is to provide in-domain samples to cover the whole training domain, and the second is to provide the extrapolation while maintaining the same data distribution.
\begin{align*}
    \beta (x) = \underbrace{ \mu(x)}_{\text{In Domain Sample}} + \underbrace{\lambda*\hat{\epsilon}_{\mu}}_{\text{Out Domain extrapolation}}.
\end{align*}
The final augmented instance feature can be defined as:
\begin{equation}
    CSU(x) = \gamma (x) (\frac{x - \mu(x)}{ \sigma (x)}) + \beta (x).
\end{equation}
This plug-and-play module can be easily inserted into any current framework. We provide pseudo-code (PyTroch) in the supplementary materials.
\section{Experiments}
\subsection{Multi-domain Classification Tasks}
\begin{figure}[!h]
  \centering
  \includegraphics[width=1.\linewidth]{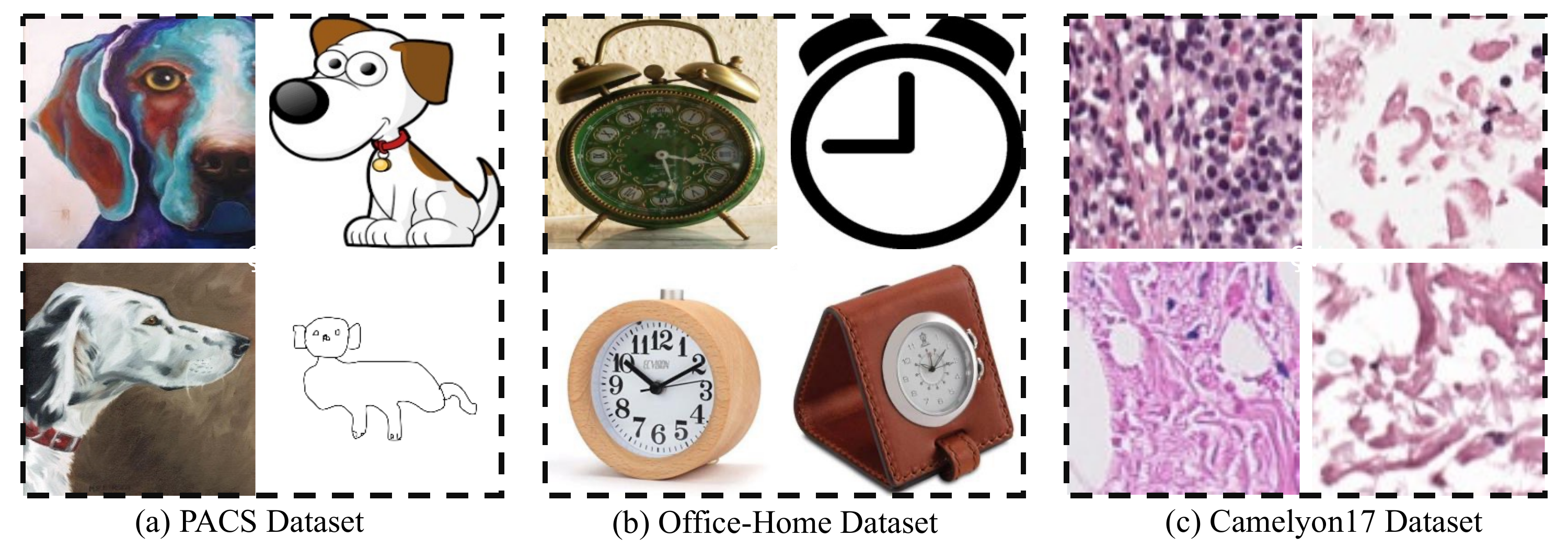}
   \caption{Some examples from multi-domain classification including (a) PACS, (b) Office-Home, and (c) Camelyon17 dataset.}
   \label{fig:dataset}
\end{figure}

We validate our model's performance on various multidomain classification tasks, including PACS, Office-Home, and Camelyon17. Figure~\ref{fig:dataset} 
 shows some examples with observable domain shifts within the same class. In all experiments, the domain tags are agnostic. Following the MixStyle, we adopt the ResNet-18~\cite{he2016deep} with ImageNet~\cite{deng2009imagenet} pre-training as the backbone for classification. We follow the Leave-One-Domain-Out strategy, which leaves one domain out for evaluation and the rest of the domains participating in the training. Adhering to a fair evaluation framework, we implement the multi-domain classification DASSL setup widely embraced in the works of Zhou et al.~\cite{zhou2022domainreview} for comparison. The batch size is set as 64. We conduct all the experiments on 2 NVIDIA A6000 GPU based on PyTorch~\cite{paszke2019pytorch} framework. 
\subsubsection{PACS classification}
\vspace{-2mm}
\quad PACS~\cite{li2017deeperPACS} is a widely used benchmark dataset for DG, which contains four domains: Photo (1,670 images), Art Painting (2,048 images), Cartoons (2,344 images), and Sketches (3,929 images). Each domain consists of seven categories for classification tasks. These domain shifts are highly suitable for validating the effectiveness of the DG algorithms. Here, we compare our model's performance with other SOTA methods, and all the evaluation metrics indicate the reported value by default.

As detailed in Table~\ref{tab: PACS classification}, our experimental outcomes reveal a significant improvement over other methodologies, underscoring the robustness of the CSU. We observe an increase of 14.3\%, 5.6\%, and 12.2\% over the baseline in Art, Cartoon, and Sketch domains, respectively. Overall, CSU has a nearly 7.8\% improvement in average accuracy across four domains. 
It should be also noted that  we adopted the pre-trained model on ImageNet; therefore, it would be hard to generate significant improvement over the baseline in the Photo domain (As discussed in~\cite{xu2020robustrandconv}), as this is expected. Nevertheless, we can still preserve the most dominant features by taking advantage of correlation modeling. Consequently, we achieve a performance drop of only around 0.1\% compared with other methods. To guarantee the reliability of the reported value, we conduct training stability analysis in the supplementary materials.

\begin{table}[!t]
  \centering
  \resizebox{\linewidth}{!}{%
  \begin{tabular}{ c| c| c c c c | c}
    \toprule
    Method & Reference & Art & Cartoon & Photo & Sketch & Average(\%) \\
    \hline
    Baseline & - & 74.3 & 76.7 & \bf{96.4} & 68.7 & 79.0\\
    Mixup~\cite{zhang2017mixup} & ICLR 2018 & 76.8 & 74.9 & 95.8 & 66.6 & 78.5 \\
    Manifold Mixup~\cite{verma2019manifoldmixup} & ICML 2019 & 75.6 & 70.1 & 93.5 & 65.4 & 76.2 \\
    CutMix~\cite{yun2019cutmix} & ICCV 2019 & 74.6 & 71.8 & 95.6 & 65.3 & 76.8 \\
    JiGen~\cite{carlucci2019domainJiGen} & CVPR 2019 & 79.4 & 75.3 & 96.0 & 71.6 & 80.5 \\
    RSC~\cite{huang2020RSC} & ECCV 2020 & 78.9 & 76.9 & 94.1 & 76.8 & 81.7 \\
    L2A-OT~\cite{zhou2020L2A-OT}& ECCV 2020 & 83.3 & 78.2 & 96.2 & 76.3 & 82.8 \\
    SagNet~\cite{nam2021SagNet} & CVPR 2021 & 83.6 & 77.7 & 95.5 & 76.3 & 83.3 \\
    pAdaIN~\cite{nuriel2021pAdaIN}& CVPR 2021 & 81.7 & 76.6 & 96.3 & 75.1 & 82.5 \\
    SFA-A~\cite{li2021sfa}  &  ICCV2021 & 81.2 & 77.8 & 93.9 & 73.7 & 81.7 \\
    MixStyle~\cite{zhou2021mixstyle} & ICLR 2021 & 82.3 & 79.0 & 96.3 & 73.8 & 82.8 \\
    DSU~\cite{li2022dsu} & ICLR 2022 & 83.6 & 79.6 & 95.8 & 77.6 & 84.1 \\
    \hline
    CSU (Ours) & - & \bf{85.0} & \bf{81.0} & 96.3 & \bf{78.4} & \bf{85.2} \\
    \bottomrule
  \end{tabular}}
  \caption{Experimental results on the PACS multi-domain classification task. CSU achieves around highly signifiant improvements over the baseline in Art, Cartoon, and Sketch domains, respectively, as: $14.3 \%, 5.6 \%,$ and $12.2\%$. Besides, CSU also shows superiority over other methods, which demonstrates its effectiveness. $\alpha$ is set as 0.1.
  }
  \label{tab: PACS classification}
  \vspace{-5 mm}
\end{table}
\vspace{-0.5mm}

We also notice that there are some domain generalization works are developed based on the DomainBed framework. While direct results comparison is not feasible, we still provide some results discussion here. Among these works, like~\cite{xu2021fourier}, the author introduced amplitude mix which linearly interpolates between the amplitude spectrum of two images to achieve data augmentation. The author achieved one classification accuracy of 84.51\% with ResNet18 which is lower than our performance, while the baseline method achieved one accuracy of 79.9\% which is higher than our performance. ~\cite{kang2022stylen} achieve one classification accuracy of 84.5\% with ResNet18 by adaptively synthesizing diverse style information with a greedy algorithm while noting that the baseline model performance is also 79.9\%~\cite{zhang2022exact} proposes to perform exact feature distribution matching in the image feature space and achieve one classification performance of 83.9 with ResNet18 while the baseline method achieves one performance of 79.5\% \cite{cha2022domain} shows that the effect of the pre-trained model with mutual-information regularization can improve the out-of-distribution improvement, which is orthogonal to all previous methods. \cite{yao2022pcl} propose a novel proxy-based contrastive learning method by replacing the sample-to-sample relations with proxy-to-sample relations. This method achieves one classification accuracy of 88.7\% with ResNet50, with ERM showing one performance of 85.5\%. 
\vspace{-2 mm}
\subsubsection{Office-Home classification}
\vspace{0 mm}
\quad Office-Home~\cite{venkateswara2017deepofficehome} is another benchmark dataset for DG, containing four domains: Art, Clipart, Product, and Real-World, and each domain consists of 65 categories. The dataset contains 15,500 images with an average of around 70 photos per class. Similarly, we compare our model's performance with other SOTA methods.
\vspace{-1mm}
\begin{table}[h]
  \centering
  \resizebox{\columnwidth}{!}{%
  \begin{tabular}{ c|  c c c c | c}
    \toprule
    Method & Art & Clipart & Product & Real & Average(\%) \\
    \hline
    Baseline & 58.8 & 48.3 & 74.2 & 76.2 & 64.4\\
    Mixup~\cite{zhang2017mixup} & 58.2 & 49.3 & 74.7 & 76.1 & 64.6 \\
    CrossGrad~\cite{shankar2018crossgrad} & 58.4 & 49.4 & 73.9 & 75.8 & 64.4 \\
    Manifold Mixup~\cite{verma2019manifoldmixup} & 56.2 & 46.3 & 73.6 & 75.2 & 62.8 \\
    CutMix~\cite{yun2019cutmix} & 57.9 & 48.3 & 74.5 & 75.6 & 64.1 \\
    RSC~\cite{huang2020RSC} & 58.4 & 47.9 & 71.6 & 74.5 & 63.1 \\
    L2A-OT~\cite{zhou2020L2A-OT} & 60.6 & 50.1 & 74.8 & \bf{77.0} & 65.6 \\
    MixStyle ~\cite{zhou2021mixstyle}& 58.7 & 53.4 & 74.2 & 75.9 & 65.5 \\
    DSU~\cite{li2022dsu} & 60.2 & 54.8 & 74.1 & 75.1 & 66.1 \\
    \hline
    CSU (Ours) & \bf{61.3} & \bf{54.9} & \bf{74.9} & 76.1 & \bf{66.8} \\
    \bottomrule
  \end{tabular}}
  \caption{Experimental results on Office-Home multi-domain classification task. We achieve around $4.3\%,  13.6\%, 0.9\%$ improvement over the baseline in the Art, Clipart, and Product domain, respectively. CSU consistently outperforms the other strong baseline models with considerable margins ($\alpha=0.4$)}
  \label{tab:Office-home classification}
  \vspace{-4mm}
\end{table}
\begin{table}[!t]
  \centering
  \resizebox{\columnwidth}{!}{%
  \begin{tabular}{ c|  c c c c c| c}
    \toprule
    Method & H1 & H2 & H3 & H4 & H5 & Average(\%) \\  \hline
    Baseline & 95.3 & 91.4 & 89.5 & 96.2 & 94.6 & 93.4\\
    MixStyle~\cite{zhou2021mixstyle}&96.1 & 91.2 & 93.0 & 95.0 & 92.7 & 93.6   \\
    pAdaIN~\cite{nuriel2021pAdaIN}& 96.6 & 93.0 & \bf{94.7} & 95.2 & 94.0 & 94.7 \\
    DSU~\cite{li2022dsu} &  \bf{96.8} & 93.3 & 91.7 & \bf{96.4} & 94.4 & 94.5  \\
    \hline
    CSU (Ours) & 96.7 & \bf{93.8} & 94.2 & 95.5 & \bf{95.5} & \bf{95.1}  \\
    \bottomrule
  \end{tabular}}
  \caption{Experimental results on Camelyon17 multi-domain classification task. H1-H5 represents five different hospitals. We can find that CSU outperforms other methods ($\alpha=0.3$).}
  \label{tab:Camelyon17 classification}
  \vspace{0mm}
\end{table}
\begin{table} [!h]
  \centering
  \resizebox{\linewidth}{!}{%
  \begin{tabular}{c|c| c c c | c |c c c}
    \toprule
    Model & ~ & Market & To &  Duke & ~ & Duke & To & Market  \\  \hline
    ResNet-50 & mAP & R1 & R5 & R10 & mAP & R1 & R5 & R10  \\ \hline
    Baseline & 19.3 & 35.4 & 50.4 & 56.4 & 20.4 & 45.2 & 63.6 & 70.9  \\ 
    RandomErase~\cite{zhong2020randomerase} & 14.3 & 27.8 & 42.6 & 49.1 & 16.1 & 38.5 & 56.8 & 64.5  \\ 
    DropBlock~\cite{zhou2021mixstyle} & 18.2 & 33.2 & 49.1 & 56.3 & 19.7 & 45.3 & 62.1 & 69.1  \\ 
    MixStyle~\cite{zhou2021mixstyle} & 23.8 & 42.2 & 58.8 & 64.8 & 24.1 & 51.5 & 69.4 & 76.2  \\ 
    pAdaIN~\cite{nuriel2021pAdaIN} & 22.0 & 41.4 & 56.4 & 62 & 24.1 & 52.1 & 68.8 & 75.5  \\
    DSU~\cite{li2022dsu} & 21.2 & 40.5 & 56 & 62.5 & 24.0 & 51.7 & 70.6 & 77.3  \\ \hline
    CSU (Ours) & \bf{24.5} & \bf{44.1} & \bf{60.3} & \bf{65.9} & \bf{24.4} & \bf{52.4} & \bf{71.4} & \bf{78.2}  \\
    \toprule
    OSNet & mAP & R1 & R5 & R10 & mAP & R1 & R5 & R10  \\ \hline
    Baseline & 25.9 & 44.7 & 59.6 & 65.4 & 24.0 & 52.2 & 67.5 & 74.7  \\
    RandomErase~\cite{zhong2020randomerase} & 20.5 & 36.2 & 52.3 & 59.3 & 22.4 & 49.1 & 66.1 & 73.0  \\ 
    DropBlock~\cite{zhou2021mixstyle} & 23.1 & 41.5 & 56.5 & 62.5 & 21.7 & 48.2 & 65.4 & 71.3  \\ 
    MixStyle~\cite{zhou2021mixstyle} & 27.2 & 48.2 & 62.7 & 68.4 & 27.8 & 58.1 & 74.0 & 81.0  \\ 
    pAdaIN~\cite{nuriel2021pAdaIN} & 28.3 & 48.8 & 62.7 & 68.1 & 27.6 & 57.5 & 74.2 & 80.3  \\
    DSU~\cite{li2022dsu} & 29.0 & 51.0 & 65.0 & 70.4 & 26.1 & 57.2 & 74.6 & 80.7  \\ \hline
    CSU (Ours) & \bf{31.1} & \bf{53.1} & \bf{67.9} & \bf{76.3} & \bf{29.8} & \bf{60.1} & \bf{77.3} & \bf{83.4}  \\
    \bottomrule
    \end{tabular}}
  \caption{Experimental results on the Duke-Market1501 Instance Retrieval Datasets. CSU achieves around $26.9\%,  19.6\%$ advancement over the baseline in mAP value using the ResNet-50 model in the Market1501 to Duke and the Duke to Market1501 experiment, correspondingly. Likewise, CSU achieves around $20.1\%,  24.2\%$ improvement for the OSNet model experiment. We could also observe similar advancements in ranking accuracy, and CSU achieves impressive improvement over other methods. }
  \label{tab: reID}
  \vspace{-6mm}
\end{table}

As shown in Table~\ref{tab:Office-home classification}, CSU achieves around $4.3\%,  13.6\%, 0.9\%$ improvement over the baseline in Art, Clipart, and Product domain, respectively. On average, CSU shows $3.7\%$ improvement over the baseline across four domains.  On the other hand, improving the Real-world images in the PACS dataset is hard due to the same reason given before. Despite this difficulty, CSU remains with a strong performance with only $0.1\%$  drop. 
\vspace{-4mm}
\subsubsection{Camelyon17 classification}
\vspace{-2 mm}
\quad Medical image analysis often suffers the most from domain shifting, given that multiple parameters, like the image acquisition device, and protocol can induce significant domain shift.  We validate the model's performance on the challenging Camelyon17 dataset~\cite{bandi2018detectioncamelyon}, containing images from five medical centers. This dataset consists of the histopathological images as input and the labels indicating whether the central region includes any tumor tissue. Due to lacking reported performance from the current literature, we conduct this experiment from scratch based on the WILDS framework proposed by Koh \cite{koh2021wilds}. Besides the baseline, we compare our model with three state-of-art strategies, including the MixStyle \cite{zhou2021mixstyle}, pAdaIN \cite{nuriel2021pAdaIN}, DSU \cite{li2022dsu}. For a fair comparison, we directly use the official implementation of each method without any modifications.

Table~\ref{tab:Camelyon17 classification} proves the effectiveness of our model. CSU achieves impressive improvement compared with the baseline or other style augmentation methods. This indicates that by taking advantage of correlation modeling, CSU can help induce a more generalized model even with extremely challenging medical data.
\vspace{-1mm}
\subsection{Instance Retrieval Experiments}
\vspace{-2 mm}
The undertaking of person re-identification, which involves the cross-camera recognition of individuals, poses a substantial domain generalization challenge. Considering each distinct camera output as its own unique domain amplifies the intricacy inherent to the person re-identification task. Following previous research, we conducted this experiment on the commonly used Duke~\cite{ristani2016performanceDuke} and Market1501~\cite{zheng2015scalablemarket} datasets. To evaluate the model's generalizability, we take one dataset as training and test the performance on the other domain. The camera data from the test domain does not participate in any training process. We adopt the exact framework implementation of MixStyle and test the CSU influence on the ResNet50~\cite{he2016deep} and 
OSNet~\cite{zhou2020L2A-OT}. Similarly, ranking accuracy and mean average precision (mAP) are performance measures. For a fair comparison, we repeat the pAdaIN and DSU experiments on the same framework with the MixStyle and use the best configuration reported in the original paper.

Table~\ref{tab: reID} shows the experiment results using two models in the two domains. We could observe that CSU outperforms other methods by a large margin. CSU achieves around $26.9\%,  19.6\%$ advancement in mAP using the ResNet-50 model in the Market1501 to Duke and the Duke to Market1501 experiment, correspondingly. Similarly, CSU achieves around $20.1\%,  24.2\%$ improvement for the OSNet model experiment. We could also observe similar advancements in ranking accuracy, and CSU achieves impressive improvement over other methods. Nevertheless, to show the effectiveness of CSU rather than position fine-tuning, we insert the permutation in all positions as described in Sec~\ref{pos_com}. The supplementary materials show that changing the inserting position can achieve even more significant performance advancement.
\subsection{Ablation Experiments}
\vspace{-2 mm}
\begin{figure*}
  \centering
  \includegraphics[width=0.8 \linewidth]{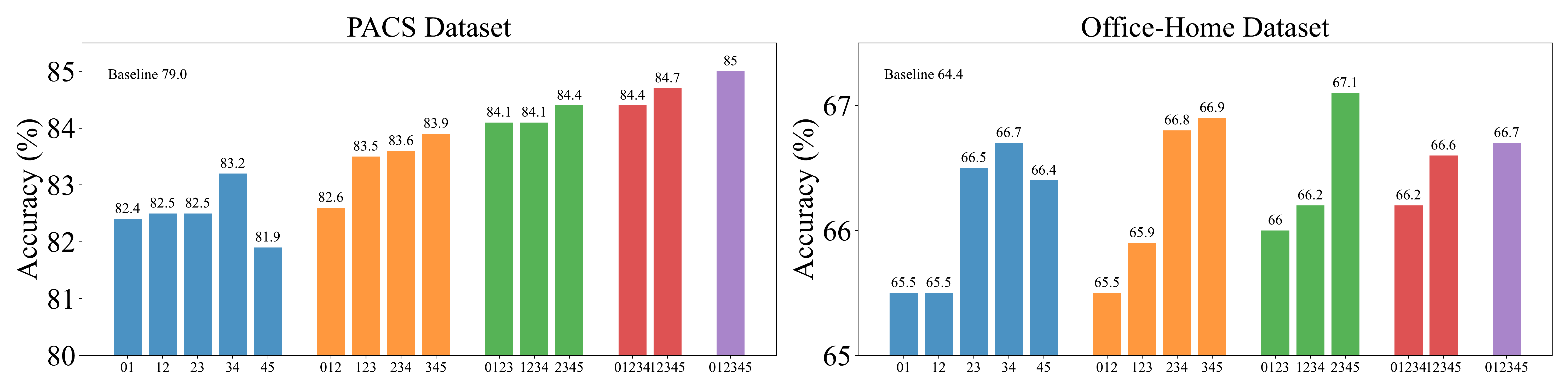}
   \caption{Influence of inserting position. Inserting 6 blocks of CSU in all potential positions achieves the best results for the PACS dataset while for the Office-Home dataset, inserting 4 blocks of CSU in the last 4 positions achieves the best results. The performance after inserting the CSU model always shows superiority over the baseline by a large margin regardless of the inserting number or position.}
   \label{fig:posselect}
   \vspace{-6 mm}
\end{figure*}
\textbf{Insert Position Selection:}
\label{pos_com}
To answer the question of where we should insert the CSU, we conduct comprehensive experiments on the PACS and Office-Home datasets using the ResNet18 structure. We investigate all possible positions of ResNet18, including the first Convolution, first Max Pooling, and 1, 2, 3, and 4 Res-block, which are named 0-5, respectively. We divide the experiment into several groups according to the inserted CSU number. Within each group, we shift the start position one by one from 0 to end. For example, for the group containing 2 CSU blocks, we will have 01, 12, 23, 34, and 45 potential combinations and five comparison experiments in total. To avoid the influence of hyperparameters, we set $\alpha=0.3$ for all experiments. Thus, we can reasonably and adequately compare the inserting position's influence on final performance.

Figure~\ref{fig:posselect} shows the ablation experiment results. Inserting 6 blocks of CSU in all potential positions achieves the best results for the PACS dataset while inserting 4 blocks of CSU in the last 4 positions achieves the best results for the Office-Home dataset. Within each group of a fixed number of CSU blocks, the performance tends to increase when we start the inserting position at the medium blocks. This trend is different with the MixStyle which the model prefers the first several blocks~\cite{zhou2021mixstyle}. We explain this phenomenon as CSU can provide more reasonable feature (statistics) augmentation due to correlation preservation. This preservation will avoid information loss in the medium or last blocks. We can also notice that compared with inserting 4, 5, and 6 blocks of CSU, inserting 2, or 3 blocks of CSU can not achieve comparable performance. This indicates that a more significant number of CSU blocks can be helpful to increase the model's generalization ability due to accumulated correlations over the blocks. It is also worth noting that no matter how we choose the inserting position, the performance of the CSU model always shows superiority over the baseline by a large margin. This firmly proves the effectiveness of the proposed model.

\begin{figure}[h]
  \centering
  \includegraphics[width=0.9 \linewidth]{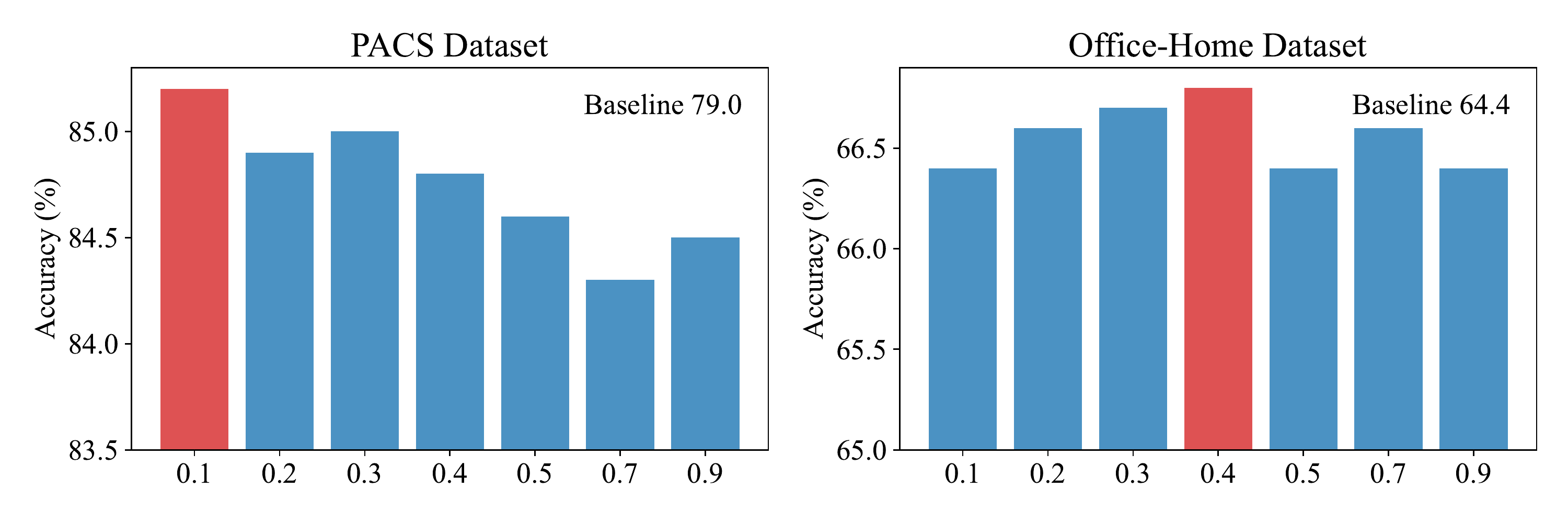}
   \caption{Influence of hyper-parameters selection. A smaller number of $\alpha<0.5$  performs better. Red indicates the best result.}
   \label{fig:hyperparam}
   \vspace{-1mm}
\end{figure}

\textbf{Hyper-parameter Selection:}
As described in the previous section, the hyper-parameter alpha determines the intensity of augmentation during training by manipulating the shape of the Beta distribution. Here, we show the influence of alpha on the PACS, Office-Home using the ResNet18 structure. Similarly, to avoid the influence of different inserting positions, we insert CSU block in all positions for every experiment. We select $\alpha$ from 0.1, 0.2, 0.3, 0.4, 0.5, 0.7, and 0.9 for one comprehensive experiment. As shown in Figure~\ref{fig:hyperparam}, we can find that a smaller number of $\alpha<0.5$ always performs better than the relatively larger number ($>0.5$). Based on these experiment results, we recommend selecting the alpha from 0.1, 0.2, 0.3, and 0.4, and the best configuration may vary according to the tasks.
\vspace{-2 mm}

\begin{figure}[h]
  \centering
  \includegraphics[width=0.9 \linewidth]{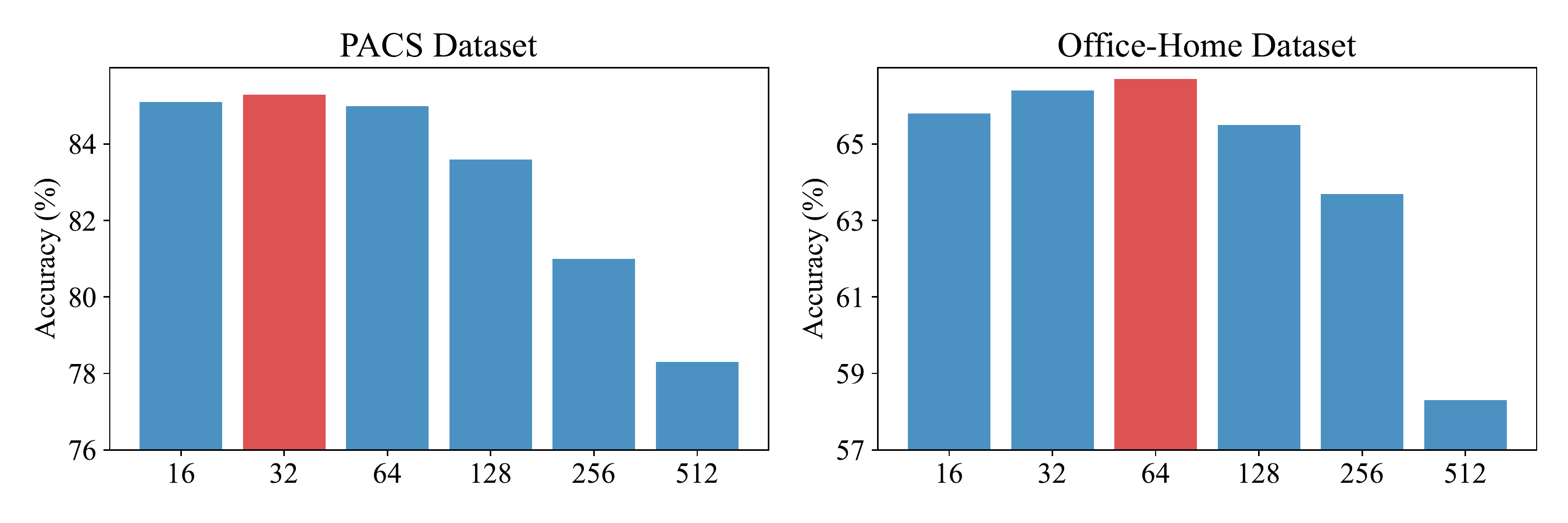}
   \caption{Effect of batch size on the two classification tasks. Too small or large batch sizes can be toxic for DG. Red indicates the best result.}
   \label{fig:batchsize}
   \vspace{-3mm}
\end{figure}
\textbf{Effect of Batch Size:}
The batch size could have a notable impact on the accuracy of correlation data, making it critical to scrutinize its effect on the final model's generalizability. Accordingly, we've introduced the CSU at every stage, as per the earlier section, and set a fixed value of $\alpha=0.3$ in all the experiments. We compare the model's performance with a batch size of 16, 32, 64, 128, 256, and 512. Figure~\ref{fig:batchsize} shows the experimental result. We found that it might be hard to estimate an accurate correlation  when the batch size is too small. Similarly, when the batch size is too large, the network tends to converge to sharp minimizers of the training and testing functions, leading to poorer generalizations, as shown in previous research~\cite{keskar2016largebatch}.

\vspace{-3 mm}

\section{Conclusion}
\vspace{-1 mm}
In this paper, we introduce the Correlated Style Uncertainty (CSU), a novel domain generalization approach that transcends linear interpolation while upholding the correlation among feature channels. 
With detailed and exhaustive ablation studies, we've determined the optimal position for integrating the CSU model, assessed the impact of sampling hyperparameters, and evaluated the ideal batch size for yielding a more universally applicable model. Comprehensive experiments across numerous datasets corroborate that CSU model notably enhances the model's ability to generalize. This research is expected to catalyze further in-depth studies on feature statistics augmentation in future.

This work is supported by NIH R01-CA246704, R01-CA240639, R15-EB030356, R03-EB032943, U01-DK127384-02S1, and U01-CA268808.

{\small
\bibliographystyle{ieee_fullname}
\bibliography{egbib}
}

\end{document}


\definecolor{codegreen}{rgb}{0,0.6,0}
\definecolor{codeblack}{rgb}{0.6,0,0}
\definecolor{codegray}{rgb}{0.5,0.5,0.5}
\definecolor{codepurple}{rgb}{0.58,0,0.82}
\definecolor{backcolour}{rgb}{1.,1.,1.}
\lstdefinestyle{mystyle}{
  backgroundcolor=\color{backcolour}, commentstyle=\color{codegreen},
  keywordstyle=\color{codeblack},
  numberstyle=\tiny\color{codegray},
  stringstyle=\color{codepurple},
  basicstyle=\ttfamily\footnotesize,
  breakatwhitespace=false,         
  breaklines=true,                 
  captionpos=b,
  frame=single,
  keepspaces=true,                  
  numbersep=5pt,                  
  showspaces=false,                
  showstringspaces=false,
  showtabs=false,                  
  tabsize=2
}
\lstset{style=mystyle}

\title{Domain Generalization with Correlated Style Uncertainty}
\maketitle

\vspace{-6mm}

The supplementary materials provide the pseudo-code for implementing the Correlated Style Uncertainty (CSU) model. To guarantee the reliability of reporting results, we also conduct training stability analysis on PACS dataset. We conduct an ablation experiment to analyze the position selection effects on the Duke-Market1501 instance retrieval task. Furthermore, we visualize the extracted feature using t-SNE projection,  which proves that CSU can help with extracting domain-invariant feature representations.
\section{Pseudo Code}
Here, we provide the pseudo-code for our CSU model. As we can observe, this code is relatively easy to implement and can be encoded into most current models. Note that we do not use the backpropagation of the normal PyTorch Eigh function for eigenvalue decomposition to avoid instability during training. This is because the gradient calculation relies on the smallest value of the eigenvalue difference $ \frac{1}{ min(\lambda_i - \lambda_j)}$~\cite{ionescu2015matrix}. 

\begin{lstlisting}[language=Python, caption=An Pytorch-like pseudo code for CSU]
# Given eps, alpha, p=0.5
# Input: x:B*C*H*W
# Output: x:B*C*H*W
def decompose(matrix):
    with torch.no_grad():
        value, vect = eigh(matrix)
    lmda = sqrt(vect.T@matrix@vect))
    return vect@lmda@vect.T

def forward(x)
    if random < p:
        return x 
    mu = mean(x, dim=(2, 3))
    sig = std(x, dim=(2, 3)) + eps
    x_norm = (x - mu) / sig 
    corr_mu = decompose(mu.T@mu)
    corr_sig = decompose(sig.T@sig)
    rand_mu = randn_like(mu)@corr_mu
    rand_sig = randn_like(sig)@corr_sig
    inten = Beta(alpha, alpha).sample(N, 1)
    mu = mu + inten*rand_mu
    sig = sig + inten*rand_sig
    x = mu + x_norm*sig
    return x
\end{lstlisting}
\label{list:algorithm}
This can induce extremely unstable training, considering that we have many zero eigenvalues, as described in the previous section. We assume that the direction is relatively stable during the training to address this issue. The key for backpropagation is calculating the eigenvalue or variance intensity for the corresponding direction. Thus, we adopt an algorithm that does not pass the gradient through the eigenvector. We show the pseudo implementation in \ref{list:algorithm}

\section{Training Reliability}
\begin{figure}[!h]
  \centering
  \includegraphics[width=1 \linewidth]{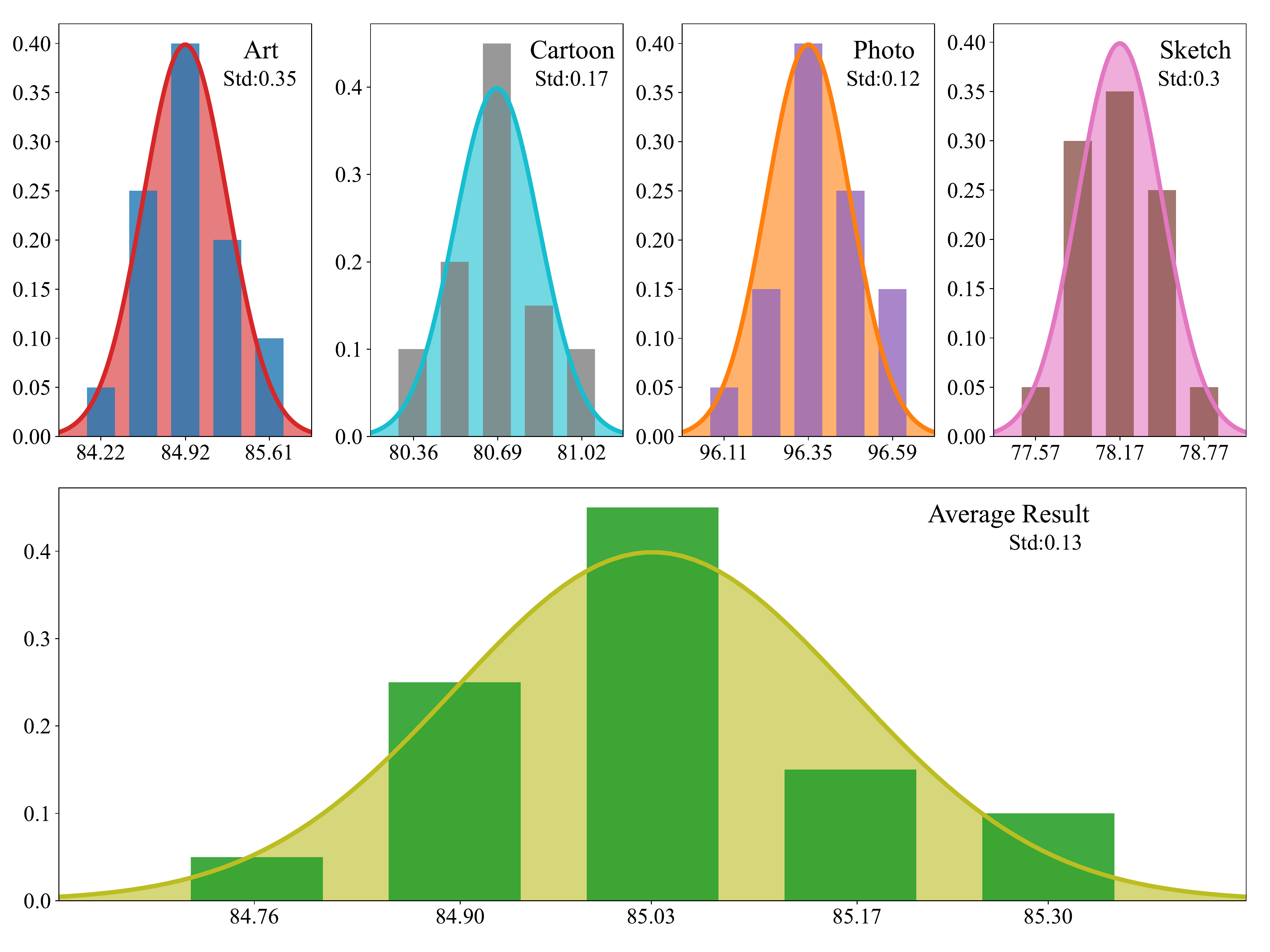}
   \caption{One visualization of training stability. We perform 20 times of experiments and calculate the standard deviations of each category and the average performance. We can observe that the training is stable given the low standard deviation value.}
   \label{fig: stability}
\end{figure}

We conduct the training process using the exact configuration on the PACS dataset multi-time. Here we set $\alpha=0.3$. We perform 20 times of experiments and calculate the performance distribution to test the training stability. The standard deviations of Art, Cartoon, Photo, and Sketch are 0.35, 0.17, 0.12, and 0.30, respectively, and the standard deviation of Average is 0.13. Figure~\ref{fig: stability} shows the result. We can observe that the standard deviation is relatively low, and the One-Sigma range is (84.90, 85.17), which indicates that the training process is consistent and reliable.
\begin{figure*}[!h]
  \centering
  \includegraphics[width=0.9 \linewidth]{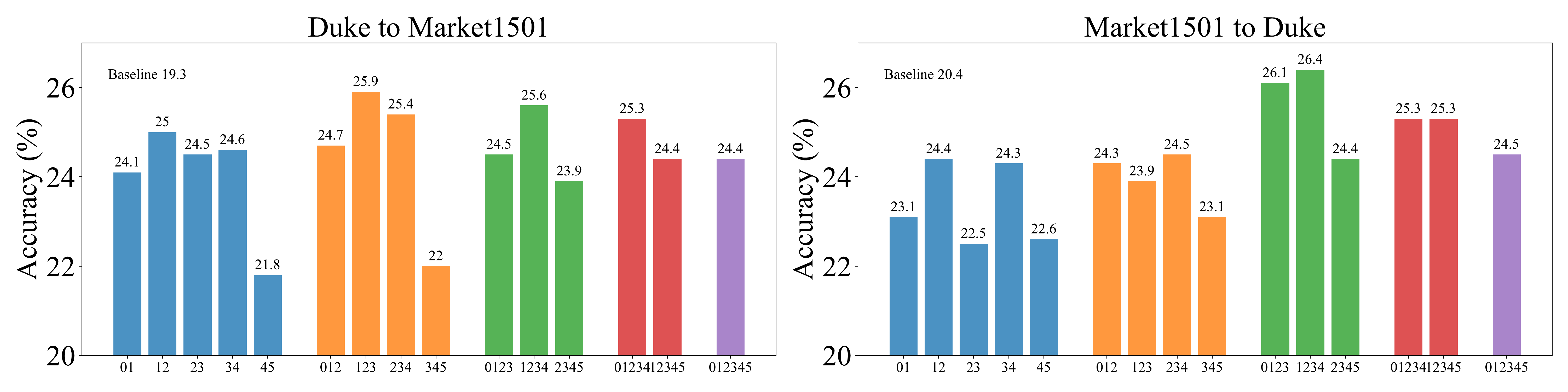}
   \caption{One visualization of different inserting positions on the instance retrieval experiments. We can achieve better performance than reporting by changing the inserting position. This shows the best position configuration might vary by task rather than one fixed conclusion.}
   \label{fig: pos ablation experiment}
   \vspace{-6mm}
\end{figure*}

\begin{figure*}[!h]
  \centering
  \includegraphics[width=.8 \linewidth]{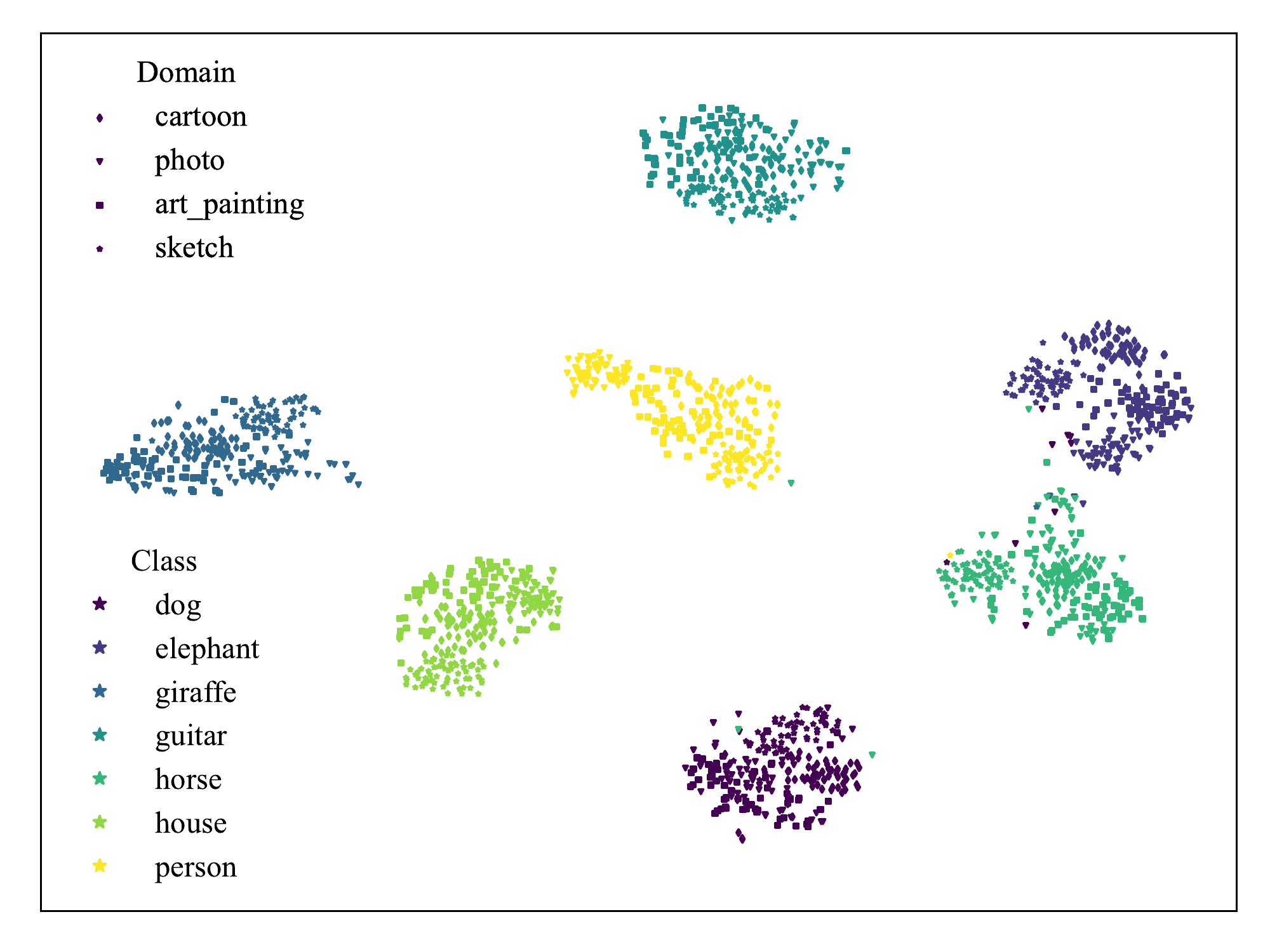}
   \caption{2-D Visualization of Flattened Feature Maps. We can clearly observe that the trained model can effectively obtain more domain-invariant feature representations. For 4 domains of the PACS dataset, including Art, Cartoon, Photo, and Sketch, we select 64 cases from every category (7 categories in total) under each domain.}
   \label{fig: feature map}
   \vspace{-5.5mm}
\end{figure*}
\section{Position Selection For Instance Retrieval}
\vspace{-2mm}
We conduct an ablation experiment to analyze the position selection effects on the Duke-Market1501 instance retrieval task. Here we show the influence of different inserting positions in Figure~\ref{fig: pos ablation experiment}. We can find that overall trends are similar to the classification tasks. Notably, we can achieve impressive improvement compared to the reported result (the "012345" group) by changing position. This indicates that the best position configuration may vary by task rather than one fixed conclusion.
\vspace{-3mm}
\section{Visualization of Flattened Feature Maps}
\vspace{-2mm}
To intuitively understand the effectiveness of our method, we provide the t-SNE visualization map of feature vectors extracted from the trained model. As shown in Figure~\ref{fig: feature map}, we can find that with the CSU, the distance between different domains within the same category is small, while the distance between different classes, regardless of the domain, is immense. Therefore, we can show that the trained model can obtain more domain-invariant feature representations, indicating a more vital generalization ability.
\vspace{-2mm}

{\small
\bibliographystyle{ieee_fullname}
\bibliography{egbib}
}